\documentclass[10pt,twocolumn,letterpaper]{article}

\usepackage{iccv}              

\newcommand{\ours}{T2MD}

\usepackage{transparent}
\usepackage{accsupp}
\newlength\savewidth\newcommand\shline{\noalign{\global\savewidth\arrayrulewidth
  \global\arrayrulewidth 1pt}\hline\noalign{\global\arrayrulewidth\savewidth}}

\usepackage{placeins}
\usepackage{pifont}
\usepackage{float}

\newcommand\blfootnote[1]{\begingroup\renewcommand\thefootnote{}\footnote{#1}\addtocounter{footnote}{-1}\endgroup}

\definecolor{iccvblue}{rgb}{0.21,0.49,0.74}
\usepackage[pagebackref,breaklinks,colorlinks,allcolors=iccvblue]{hyperref}

\def\paperID{10442} 
\def\confName{ICCV}
\def\confYear{2025}

\title{Diffusion Transformer-to-Mamba Distillation for \\ High-Resolution Image Generation}

\def\authorBlock{
    Yuan Yao$^{1,2\textsuperscript{*}}$~~~
    Yicong Hong$^2$~~~
    Difan Liu$^2$~~~
    Long Mai$^2$~~~
    Feng Liu$^2$~~~
    Jiebo Luo$^1$  \\
    $^1$~University of Rochester~~~~~$^2$~Adobe Research\\
    {\tt\small yyao39@ur.rochester.edu jluo@cs.rochester.edu} \\
    {\tt\small \{yhong, diliu, malong, fengl\}@adobe.com}
}
\author{\authorBlock}

\begin{document}
\maketitle
\begin{figure*}[t]
    \centering
    \includegraphics[width=\linewidth]{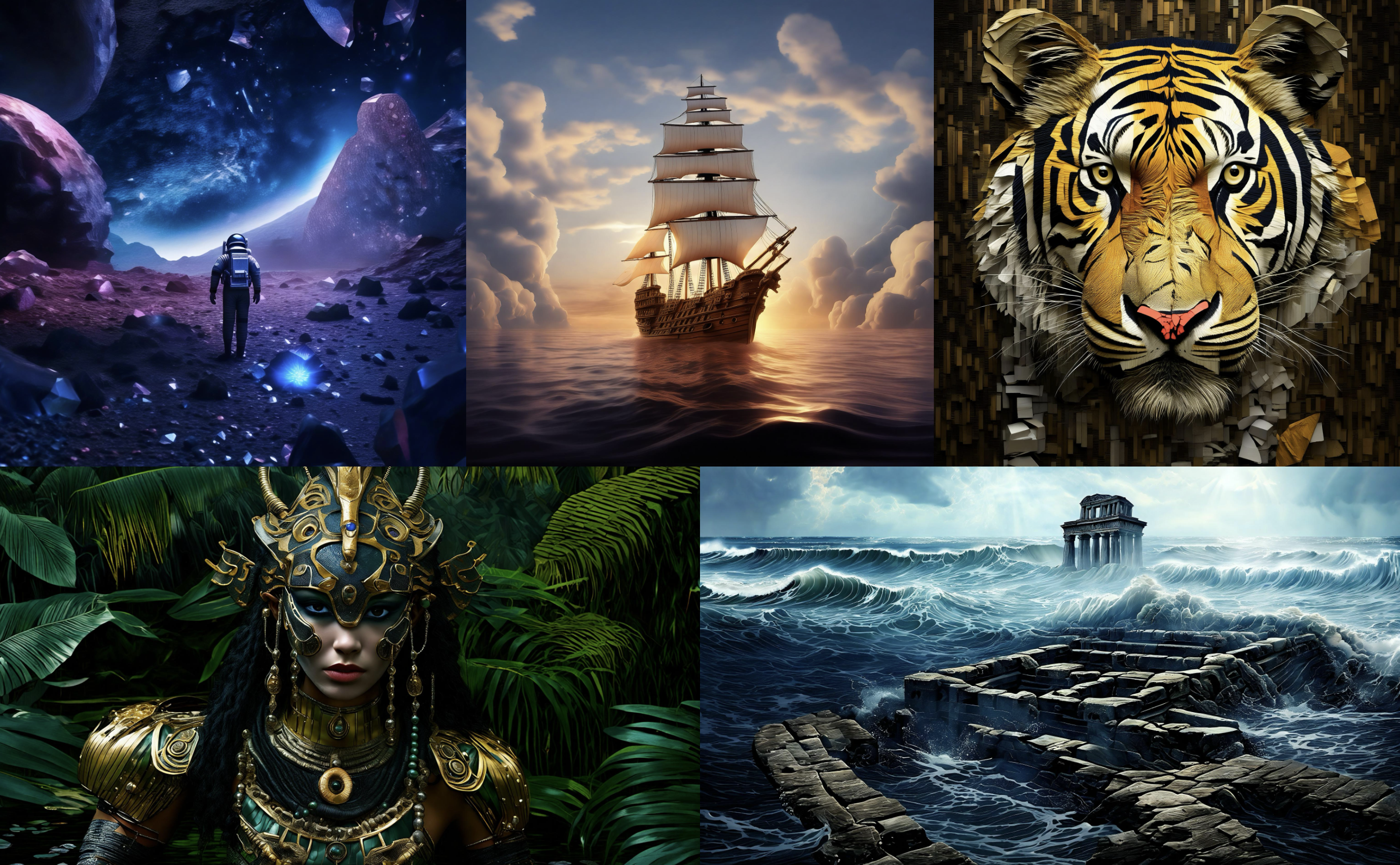}
    \caption{High-resolution generation results from our 0.7B \ours{} model. The resolutions of these images in the top and bottom rows are 2048$\times$2048 and 2688$\times$1536, respectively.}
    \label{fig:teaser}
\end{figure*}
\begin{abstract}
The quadratic computational complexity of self-attention in diffusion transformers (DiT) introduces substantial computational costs in high-resolution image generation.
While the linear-complexity Mamba model emerges as a potential alternative, direct Mamba training remains empirically challenging.
To address this issue, this paper introduces diffusion transformer-to-mamba distillation (\ours{}), forming an efficient training pipeline that facilitates the transition from the self-attention-based transformer to the linear complexity state-space model Mamba.
We establish a diffusion self-attention and Mamba hybrid model that simultaneously achieves efficiency and global dependencies.
With the proposed layer-level teacher forcing and feature-based knowledge distillation, \ours{} alleviates the training difficulty and high cost of a state space model from scratch.
Starting from the distilled 512$\times$512 resolution base model, we push the generation towards 2048$\times$2048 images via lightweight adaptation and high-resolution fine-tuning.
Experiments demonstrate that our training path leads to low overhead but high-quality text-to-image generation. 
Importantly, our results also justify the feasibility of using sequential and causal Mamba models for generating non-causal visual output, suggesting the potential for future exploration.

\end{abstract}

\section{Introduction}

\blfootnote{\textsuperscript{*} Work done during an internship at Adobe Research.}

In recent years, the diffusion transformer (DiT) model 
has become a standardized design of image generation backbones~\cite{peebles2023scalable}. 
However, using transformers introduces non-trivial computational burdens, especially as image resolution increases.
This computational cost is associated with transformers’ self-attention mechanism that scales quadratically with the number of image tokens, posing significant challenges for high-resolution image synthesis.


At the same time, linear complexity models (LCM), such as Mamba~\cite{gu2023mamba}, have demonstrated strong potential as efficient and scalable alternatives to Transformer architectures in natural language processing\cite{poli2023hyena, lieber2024jamba, zhao2024cobra, huang2024ml} and visual understanding~\cite{liu2024vmambavisualstatespace, islam2023efficient, nguyen2022s4nd, xu2024survey}. 
In the recent year, there has been a surge of Mamba-based diffusion models~\cite{hu2024zigma, mo2024scaling, teng2024dim, fei2024dimba}. These studies have demonstrated that employing the multi-directional scanning and hybrid model design (\textit{i.e.}, mixed transformer and mamba) allows mamba-based diffusion models to achieve high-fidelity image generation on low-resolution tasks.

However, the target resolution in prior Mamba-based diffusion models have been capped at the 512$\times$512 resolution. 
The main reasons behind this are that (1) a high-quality text-to-image generation model heavily relies on a well-performing base model, whose training requires an extensively large amount of computational resources, and it has been shown that Mamba is harder to converge than Transformer models in general~\cite{waleffe2024empirical}. 
(2) On the other hand, it remains to be explored whether a sequential and casual state-space model is capable of modeling the long-range non-causal dependency among the visual tokens in large images.
In light of this, we seek to answer the following question: \textit{Can we find an efficient and effective training path for the Mamba-based models for high-resolution text-to-image generation?}

Inspired by the Transformer-to-LCM distillation literature in natural language processing~\cite{mercat2024lll, zhang2024hedgehog}, we propose to first establish a Mamba diffusion model by distilling a well-performing teacher DiT to bypass the expensive base model training (\textit{abbrev.} \ours{}).
However, unlike causal-to-causal distillation in NLP tasks, the formulation between non-causal attention and causal Mamba can lead to accumulated errors in the distillation process.
To solve this issue, we propose the layer-level teacher forcing, which explicitly aligns intermediate outputs of student Mamba layers by injecting pseudo-ground-truth input features from each teacher self-attention layer.
We then perform knowledge distillation to train the token mixers (\ie, self-attention and Mamba layers), followed by a model adaptation phase.
Lastly, high-resolution fine-tuning further improves image generation, starting with 512$\times$512 resolution and gradually advancing to 2048$\times$2048 resolution. 
This structured approach enables effective knowledge transfer and scaling to high resolutions, improving the model’s robustness and image quality.

We construct a 0.7B hybrid diffusion Mamba model comprised of 86\% of Mamba blocks and 14\% of self-attention blocks for efficient computation and preserving global interactions, respectively.
Experiments demonstrate that \ours{} greatly boosts the performance of the diffusion Mamba model by 0.28 in GenEval score, and can reach the teacher model in quality.
The hybrid Mamba model is efficient in generating high-resolution images. 
Compared to the DiT models, \ours{} is 
2.1$\times$ faster in generating 4K images in terms of sampling image throughput.


In summary, our contributions are threefold:
\begin{itemize}
    \item We propose a hybrid diffusion Mamba model for efficient high-resolution image generation. Our model is the first SSM-based model 
    to achieve beyond-2k text-to-image generation.
    \item We design an effective and efficient knowledge distillation scheme for training the diffusion Mamba model, using layer-level teacher forcing to mitigate the intrinsic difference between non-causal self-attention and causal Mambas.
    \item We conduct extensive experiments to validate the effectiveness of our method, suggesting the feasibility of using sequential and causal Mamba models for generating non-causal visual output.
\end{itemize}


\section{Related work}

\subsection{Diffusion-based image generation}
Diffusion models~\cite{ho2020denoising} have been prevailing in image generation tasks. Early diffusion image generation models rely on a U-Net~\cite{ronneberger2015unet} internal architectures, characterized by their down-sampling and up-sampling blocks with skip connections~\cite{dhariwal2021diffusion, rombach2022high, podell2023sdxl, saharia2022imagen, ho2021cascaded, nichol2022glide}. 
However, models with U-Net structure are not able to easily scale up, thus lacking the capability for high-resolution generation. The DiT architecture is therefore proposed to solve this issue by replacing the convolution-based model with an architecture composed solely of attention~\cite{vaswani2017transformer} and MLP layers~\cite{peebles2023scalable}. Such methods have been adopted in a number of works, and have shown advantage in terms of scalability~\cite{chen2023pixartalpha, gao2024lumina-next, gao2024lumin-t2x, chen2023gentron, gao2023masked, zheng2024maskdit} and multi-modal learning~\cite{bao2023one, chen2023palixscalingmultilingualvision}. With the success of Sora~\cite{videoworldsimulators2024} on high resolution text-based video generation, diffusion Transformers gradually become the core backbone for most of the current state-of-the-art media generation models. While DiT is capable of large-scale training, the quadratic complexity has made it computationally inefficient to train under ultra high resolution. In our work, we propose to further resolve this issue by replacing the attention layers with mamba to reduce the computational cost.

\subsection{State space models}
State space models (SSMs)~\cite{hamilton1994state} are a class of models sharing the state transfer equation. The original structured SSMs (S4) are introduced in~\cite{gu2021efficiently}. Many variants of SSMs have also been proposed~\cite{smith2024convolutional, peng2023rwkv, fu2023simple, fu2022hungry, mehta2022long}. 
Among all the variants, Mamba stands out as a time-varying selective SSM and has shown superior potential for modeling long sequences~\cite{gu2023mamba, dao2024transformers}. Mamba has applications in various fields, including natural language processing~\cite{poli2023hyena, lieber2024jamba, zhao2024cobra, huang2024ml} and visual understanding~\cite{liu2024vmambavisualstatespace, islam2023efficient, nguyen2022s4nd, xu2024survey, li2024videomamba}. Despite its success in these tasks, it has been shown that Mamba is not capable of long-range dependencies~\cite{waleffe2024empirical}. Therefore, we propose the hybrid Mamba model that includes some attention layers to solve this issue. Recently, a number of works have been working on integrating Mamba into diffusion models~\cite{hu2024zigma, fei2024dimba, teng2024dim, mo2024scaling, gao2024matten}. While these works show promising results, none of these works succeed in direct generation over 512$\times$512 resolution. In this work, we propose to train a diffusion Mamba model for image generation.

\subsection{Knowledge distillation}
Knowledge distillation (KD) methods train a student model under the supervision of a teacher model~\cite{hinton2015distilling} and are commonly used for model compression~\cite{gou2021knowledge, jha2024justchop, xia2023sheared}. Current KD approaches include response-based~\cite{zhou2021rethinking,hinton2015distilling}, relation-based~\cite{tian2019contrastive,tung2019similarity}, and feature-based methods~\cite{jin2019knowledge,heo2019comprehensive,chen2021cross}. Our approach falls under feature-based KD, aiming to supervise student representation learning through intermediate representations so that the student network can mimic the teacher behavior. Recently, a number of works have tried to distill across architectures, including distilling Transformer weights into linear transformers~\cite{liu2024linfusion, kasai2021distill, mercat2024lll, zhang2024hedgehog}. However, the method of distilling non-causal transformer into state space models remains unexplored. Inspired by previous works, we propose to distill DiTs into diffusion Mamba to save training resources.
\section{Method}


\begin{figure}[t]
    \centering
    \includegraphics[width=\linewidth]{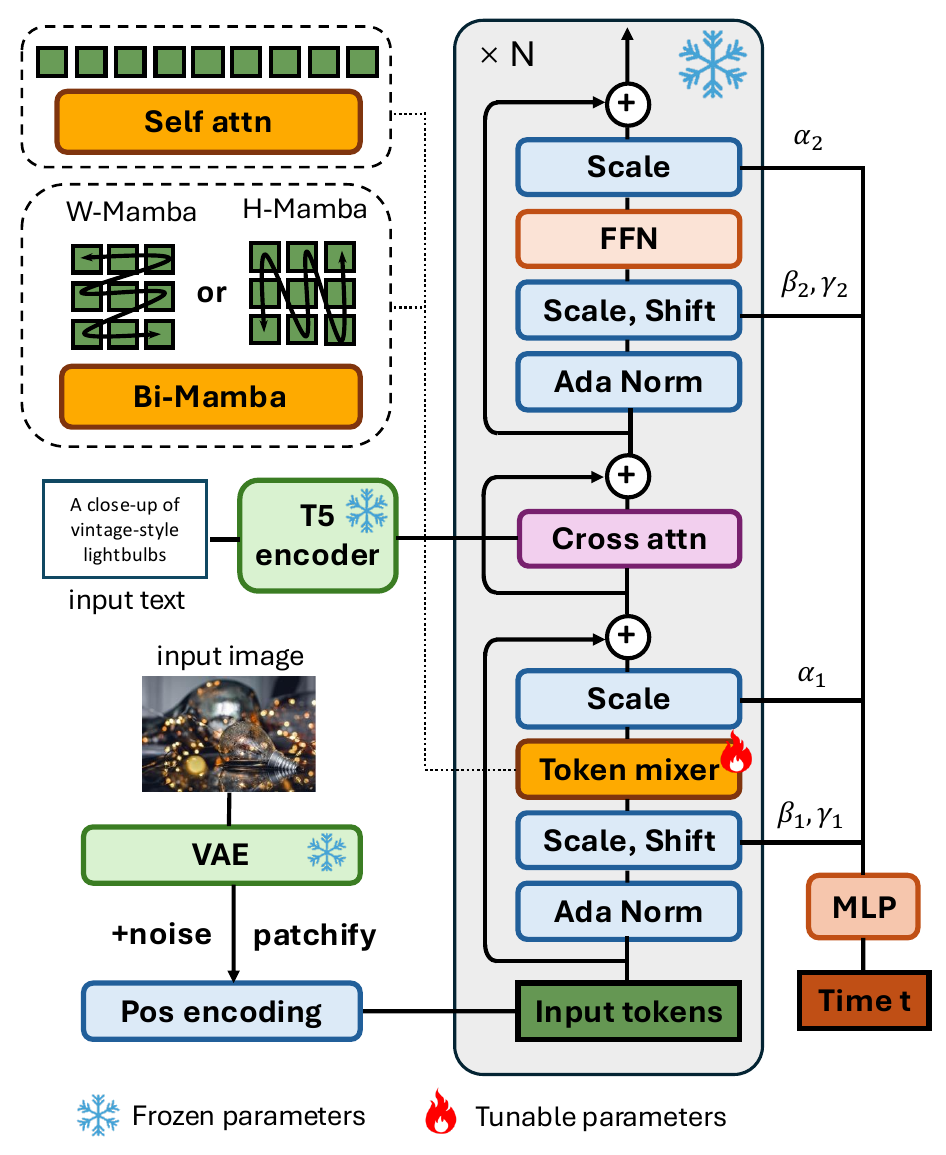}
    \caption{The architecture of our hybrid Mamba model. During the knowledge distillation stage, all the parameters except the token mixers are frozen.}
    \vspace{-10px}
    \label{fig:architecture}
    \vspace{-5px}
\end{figure}

\subsection{Motivation}

Considering the computational efficiency at inference, applying a diffusion Mamba model can be a promising direction for high resolution image generation. 
However, training high-resolution diffusion models typically involves first developing a strong base model at a lower resolution, which requires substantial computational resources. Recent research has also shown that Mamba models generally underperform compared to transformers and are harder to converge~\cite{waleffe2024empirical}. Therefore, training a Mamba-based diffusion model from scratch is likely inefficient for this purpose. 

On the other hand, state space duality~\cite{dao2024transformers} provides the theoretical soundness to train diffusion Mamba layers to mimic the diffusion Transformer layers.
Given the success of current DiT text-to-image generation models, we propose to distill the pre-trained DiT models into our Mamba-based diffusion model.
Specifically, we choose the open-source PixArt-$\alpha$~\cite{chen2023pixartalpha} as the teacher model due to its transformer-dense architecture, appropriate scale and preeminent text-to-image performance.

The advantage of such knowledge distillation can be concluded into two aspects. First, we obtain a well-trained Diffusion Mamba model with relatively few computational resources. 
Second, we extend the high-resolution generation efficiencies of current DiT models by replacing transformer layers with Mamba layers. 

\subsection{Preliminary}
\subsubsection{Diffusion Model}
The diffusion process is central to our model's capability for generating detailed and coherent images. It begins with a forward process that gradually introduces Gaussian noise into a clean image, transitioning it to a purely noisy state across several pre-defined timesteps. This process is mathematically defined as:
\begin{equation}
    q(x_t | x_{t-1}) = \mathcal{N}(x_t; \sqrt{1-\beta_t} x_{t-1}, \beta_t \mathbf{I}),
\end{equation}
where $\beta_t$ represents the variance schedule of the noise added at each step.

The reverse process, a denoising diffusion probabilistic model, learns to reconstruct the original image by sequentially reducing the noise. This is achieved through a parameterized model $\theta$ that predicts the noise in the image, allowing for the recovery of the clean image from the noisy input:
\begin{equation}
    p_\theta(x_{t-1} | x_t) = \mathcal{N}(x_{t-1}; \mu_\theta(x_t, t), \Sigma_\theta(x_t, t)),
\end{equation}
where $\mu_\theta$ and $\Sigma_\theta$ are learned functions of the model.

\subsubsection{Mamba}
Mamba~\cite{gu2023mamba} originates from state space model (SSM). SSMs take 1-D sequence $x \in \mathbb R^T$ as input and outputs the sequence $y \in \mathbb R^T$ of the same length. The representation is modeled by the hidden state $h$ in SSMs. The transfer between states can be formulated as 
\begin{equation}
\begin{aligned}
    h_t & = \mathbf{A}_th_{t-1} + \mathbf{B}_tx_t, \\
    y_t & = \mathbf C_t^T h_t,
\end{aligned}
\end{equation}
where $\mathbf{A} \in \mathbb R^{(T, N, N)}, \mathbf{B} \in \mathbb R^{(T, N)}$, and $\mathbf{C} \in \mathbb{R}^{(T,N)}$ are the parameters.

Mamba optimizes the efficiency of traditional SSM by relaxing the parameters $\mathbf A, \mathbf B$ and $\mathbf C$ to time-invariant. Mamba 2 further reduces $\mathbf A$ to be diagonal with identity values~\cite{dao2024transformers}. Mamba employs a discrete approximation for the continuous state dynamics, controlled by a parameter $\Delta$. The general formulation of SSM is thus rewritten as
\begin{equation}
\begin{aligned}
    \mathbf{\bar A} &= \exp(\Delta \mathbf{A}),~ \mathbf{\bar B} = (\Delta \mathbf A)^{-1}\exp(\Delta \mathbf{A} - \mathbf{I}) \cdot \Delta \mathbf{B}, \\
    h_t &= \mathbf{\bar A} h_{i-1} + \mathbf{\bar B} x_i,~~ y_i = \mathbf{C} h_i.
\end{aligned}    
\end{equation}

With these design choices, the Mamba model largely enhances GPU efficiency by utilizing parallel scanning. With global convolution kernels, Mamba achieves linear computational complexity.

While Mamba shows advantage with linear computational efficiency, the formulation of SSMs can lead to the issues of long-term forgetting and in-context learning difficulties. 
Nevertheless, research has demonstrated that these issues can be solved by mixing a small proportion of self-attention layers into Mamba models, and the hybrid model can even show greater capabilities compared with pure self-attention models~\cite{waleffe2024empirical}.
Therefore, in this paper, we propose the hybrid diffusion model that consists of a mixture of self-attention layers and Mamba layers.

\subsection{Model architecture}
\paragraph{Teacher model.}
To explore the feasibility and effectiveness of diffusion Transformer-to-Mamba knowledge distillation, this paper chooses PixArt-$\alpha$~\cite{chen2023pixartalpha}, a Transformer-dense DiT model, as the teacher model. PixArt-$\alpha$ has the simple-and-general model design, appropriate model scale and superior generation quality, thus suitable for validating the distillation process.

\paragraph{Our framework.} The overall framework of the proposed hybrid diffusion Mamba model can be found in Fig.~\ref{fig:architecture}. We establish a diffusion self-attention and Mamba hybrid model that simultaneously achieves efficiency and global dependencies. The variational autoencoder $(\mathcal{E}, \mathcal{D})$ is employed to project the image into latent space and decode the \ours{} model's output. The noisy latent is split into patches, and the positional encodings are added, followed by flattening the tokens and feeding into the model blocks. 
Consistent with the teacher model, our entire model consists of $28$ blocks in total. The main components in each block are the token mixer, the cross-attention layer, and the feed-forward network.

\paragraph{Token mixer.}  
The token mixer is either a self-attention layer or a Mamba layer.
Specifically, we employ Mamba 2~\cite{dao2024transformers} due to its superior efficiency. As Mamba naturally operates on 1D sequences, we first transform the 2D image tokens into 1D sequences through a raster scan. To ensure Mamba captures non-causal global context, we adopt a bidirectional scanning strategy.
The scanning directions are width-first and height-first interleaved.
The weights of the Mamba model are shared between these two scans, and their outputs are fused by a linear projection layer.
In our proposed model, we configure the architecture to have 4 self-attention blocks and 24 Mamba blocks.

\paragraph{Text and timestep conditioning.}
The text input is tokenized by a T5 encoder~\cite{raffel2020exploringt5} and then fed into the cross-attention layer.
The timestep is injected into the model through adaptive layer normalization~\cite{peebles2023scalable}, with the scale and shift parameters $\alpha, \beta$ and $\gamma$ learned by MLP layers. 

\subsection{Multi-stage training strategy}

\begin{figure}[t]
    \centering
    \includegraphics[width=\linewidth]{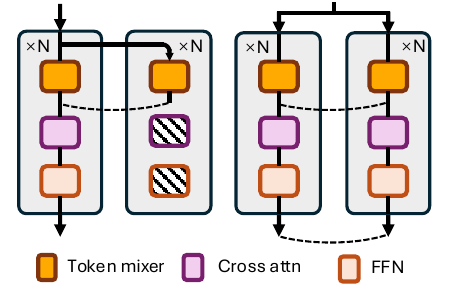}
    \vspace{-10px}
    \caption{Illustration of the teacher-student alignment in layer-level knowledge distillation \textit{(left)} and feature-based knowledge distillation \textit{(right)}.}
    \label{fig:pipeline}
    \vspace{-5px}
\end{figure}

\subsubsection{Layer-level teacher forcing}
\label{sec:forcing}
Since Mamba models are more difficult to train~\cite{waleffe2024empirical}, we aim to distill the pre-trained teacher DiT weights into diffusion Mamba models.
One common way of knowledge distillation is to minimize discrepancies in output features between each corresponding layer from the teacher and student model. 
However, applying this approach to distill knowledge from a non-causal self-attention model to a causal Mamba model introduces a fundamental challenge.
Specifically, the intrinsic difference in their architectures causes minor deviations in early student layers to accumulate and amplify throughout deeper layers, leading to severe error propagation and ultimately destabilizing the training of the Mamba model. 

To overcome this challenge, we propose the layer-level teacher-forcing strategy tailored explicitly for distilling self-attention into Mamba. Unlike conventional distillation approaches, which rely solely on student-generated intermediate representations, our method leverages intermediate inputs from teacher’s self-attention layers as pseudo-ground-truth inputs for each student Mamba layer.
By explicitly aligning layer-wise inputs with those of the teacher, our strategy effectively mitigates error propagation and enhances the consistency and quality of knowledge transfer between fundamentally different architectures.
Formally, the input for teacher token mixer at layer $n$ can be written as
\begin{align}
    \mathbf{h}_{\theta^{'}}^{(n)} = \epsilon^{(n)}_{\theta^{'}}(z_t, t, \tau),
\end{align}
where we denote the teacher model as $\epsilon_{\theta'}$, the timestep as $t$, the text prompt as $\tau$, and the input latent the noisy input latent as $z_t$. 
The teacher forcing loss is then formulated as
\begin{align}
    \mathcal{L}_{\text{forcing}} = \sum_{n=1}^N \left\| \mathbb{I}(n) \left( \operatorname{MA}_{\theta}^{(n)}(\mathbf{h}^{(n)}_{\theta^{'}}) - \operatorname{SA}_{\theta^{'}}^{(n)}(\mathbf{h}^{(n)}_{\theta^{'}}) \right) \right\|_2^2,
\end{align}
where $\epsilon_\theta$ is the student model, $\mathbb{I}(n)$ is an indicator function that the n-th student block is a Mamba block, and $\operatorname{MA_{\theta}^{(n)}}, \operatorname{SA}_{\theta^{'}}^{(n)}$ denote the student token mixer Mamba and teacher token mixer self-attention, respectively. 

The training objective is to optimize the Mamba layers in the student model by minimizing the teacher forcing loss within diffusion process, formulated as
\begin{align}
    \bar{\theta} = \arg \min_{\bar{\theta}}\mathbb{E}_{q(z_t | z_0)} \mathcal L_{\text{forcing}}.
\end{align}
This training paradigm prevents early-layer deviations from snowballing into larger discrepancies in deeper layers. The student model can better approximate the hierarchical representations of the teacher, resulting in superior distilled performance.

\subsubsection{Knowledge distillation} \label{sec:kd}
In this stage, we conduct knowledge distillation. We inherit the Mamba weights from the teacher forcing step, and initialize all other model parameters by copying the weights from the teacher model.
During knowledge distillation, the parameters of the token mixers (either self-attention layers or mamba layers) are unfrozen, while all other components are frozen.

Our distillation loss consists of three components: the MSE loss, the pseudo loss, and the token mixer loss. 
The first loss term is the standard diffusion MSE loss, formulated as 
\begin{align}
    \mathcal{L}_{\text{mse}} = \left\| \epsilon - \epsilon_{\theta}(z_t, t, \tau) \right\|_2^2.
\end{align}
The pseudo loss is the loss between the predicted noise from the student model and the soft labels generated by the teacher model, which can be formulated as
\begin{align}
    \mathcal{L}_{\text{pseudo}} = \left\| \epsilon_{\theta^{'}}(z_t, t, \tau)- \epsilon_{\theta}(z_t, t, \tau) \right\|_2^2.
\end{align}

For the token mixer loss, we compare the output of each corresponding token mixer from the teacher model and the student model. The token mixer loss can be formulated as
\begin{align}
    \mathcal{L}_{\text{mixer}} = \frac{1}{N}\sum_{n=1}^N \left\| \epsilon^{[n]}_{\theta^{'}}(z_t, t, \tau)- \epsilon^{[n]}_{\theta}(z_t, t, \tau) \right\|_2^2,
\end{align}
where $N$ is the number of layers.
The distillation loss for the knowledge distillation stage can be formulated as 
\begin{align}
    \mathcal{L}_{\text{distill}} = \mathcal{L}_{\text{mse}} + \lambda_1 \cdot \mathcal{L}_{\text{pseudo}} + \lambda_2 \cdot \mathcal{L}_{\text{mixer}},
\end{align}
where $\lambda_1$ and $\lambda_2$ are hyper-parameters.
The training objective is to minimize this loss within diffusion process, formulated as
\begin{align}
    \theta = \arg \min_{\theta}\mathbb{E}_{q(z_t | z_0)} \mathcal L_{\text{distill}}.
\end{align}


\subsubsection{Model adaptation}
\label{sec:adapt}
Since the teacher model is not capable of high-resolution image generation, some components of the teacher model may not be adequately adapted for the generation of high-resolution images. 
For instance, the commonly adopted sine-cosine positional encoding is not suitable for multi-resolution training. Although the positional interpolation trick~\cite{xie2023difffit} can enhance the adaptation performance at a larger resolution, it still fails to achieve zero-shot higher-resolution sampling.
To demonstrate the flexibility of the distilled model, we propose the optional model adaptation stage to replace model components, such as the positional embeddings, VAE encoder and T5 encoder.
We replace the sine-cosine positional embeddings with a centered sine-cosine positional embedding normalized by the length of the long edge to enable zero-shot high-res sampling. 
Besides, the SD v1.5 VAE adopted by the PixArt-$\alpha$ model is not suitable for high-resolution generation~\cite{chen2024pixart}.
We also replace the original VAE from the teacher model with SDXL's VAE~\cite{podell2023sdxl}.
Although many parts of the model have been completely changed, the model can converge within 100k steps during adaptation.

\subsubsection{High resolution finetuning}\label{sec:ft}
In the previous stages, the model is only trained with lower-resolution images, whereas finetuning on high-resolution data is necessary for high-resolution image generation. We fine-tune the distilled 512$\times$512 resolution model with two stages to reach 2048$\times$2048 resolution. In the first stage, we fine-tune our model with mixed 512$\times$512 data and 1024$\times$1024 data for 40k steps. The ratio for 1024$\times$1024 data is 80\%. In the second stage, we fine-tune our model on 2048$\times$2048 data for 20k steps. After fine-tuning, the model can generate high-fidelity 2048$\times$2048 images and is also capable of zero-shot 4K generation. 
\section{Experiments}

\subsection{Implementation details}

The \ours{} model uses the pre-trained VAE from SDXL~\cite{podell2023sdxl} and the pre-trained Flan-XXL T5 encoder~\cite{chung2024scalingt5}. The VAE and T5 encoder are frozen throughout the training.
The \ours{} model contains $28$ core blocks. 
The self-attention blocks, H-Mamba blocks and W-Mamba blocks are interleaved by ``SA-(HM-WM)$\times$3-SA-(HM-WM)$\times$3-SA-(HM-WM)$\times$3-SA-(HM-WM)$\times$3". 
For the Mamba 2 layer, we choose the SSM state dimension of 256 and the SSM expand factor of 2. 
The hidden dimension of the model is 1152. The patch size of our model is $2$. We choose to use the $\varepsilon$ prediction with quadratic noise schedule. We keep an exponential moving average with decay rate $0.9999$. 
During knowledge distillation, the hyperparameters are set to $\lambda_1 = 0.5, \lambda_2 = 0.2$.
The training dataset comprises 200M image-text pairs.

\begin{figure}[t]
    \centering
    \vspace{15px}
    \includegraphics[width=1.0\linewidth]{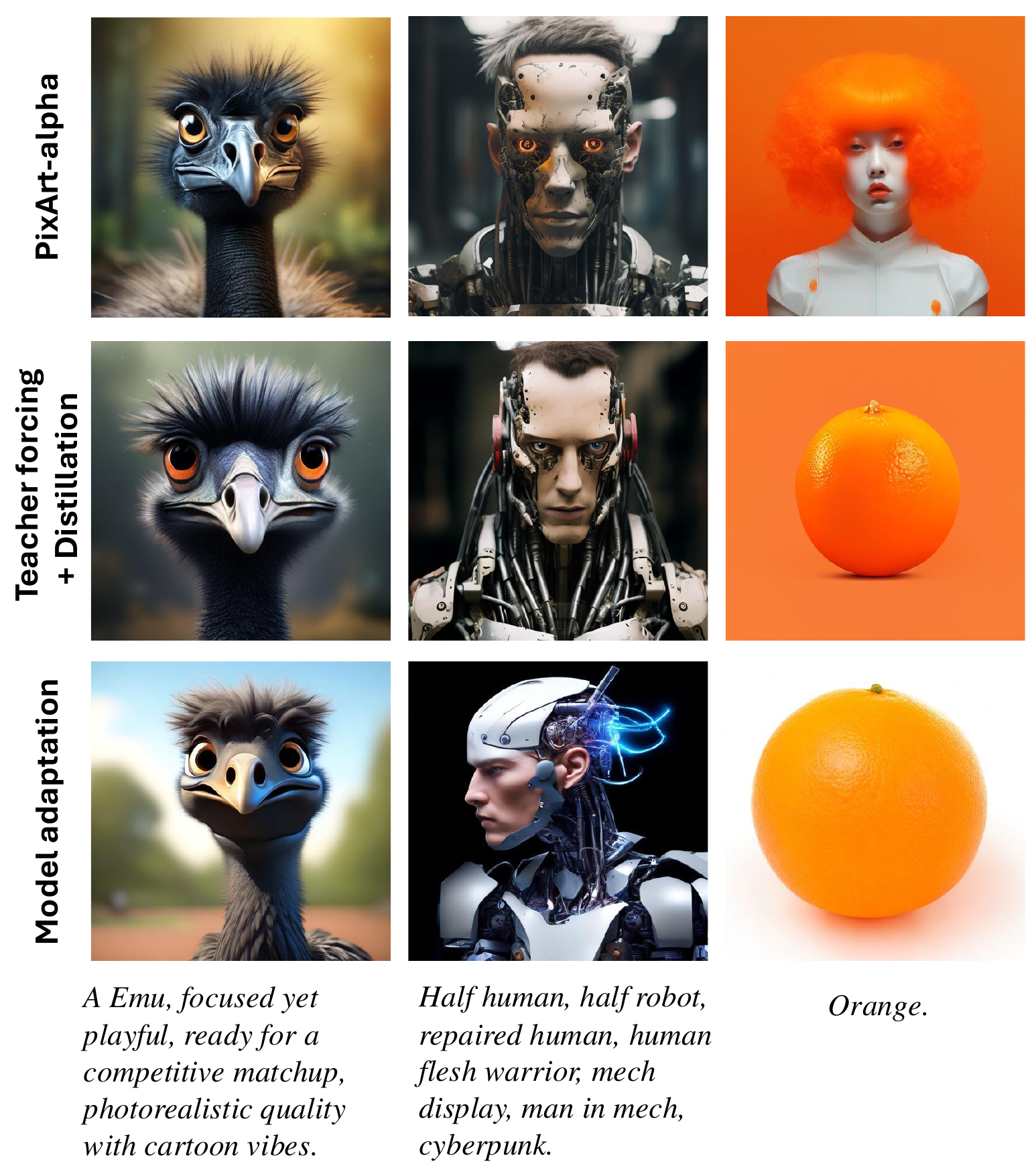}
    \caption{Results for each 512$\times$512 stage. During the teacher forcing stage and knowledge distillation stage, where we only train the token mixers, most of the image details have been preserved. The \textit{orange} case demonstrates that the Mamba layers are capable of learning the semantics during the training process.
    }
    \label{fig:512}
\end{figure}

\subsection{Ablation study}
\subsubsection{Knowledge distillation analysis}
In this section, we study the effectiveness of the proposed knowledge distillation framework. We assessed the performance of our image generation model using the GenEval score~\cite{ghosh2023geneval}.

The results are presented in Tab.~\ref{tab:ablation}. All the ablation results are presented in the second section of the table. "Baseline hybrid Mamba`` refers to the baseline method of directly training the model hybrid Mamba model we propose for 200k steps. As demonstrated by the GenEval results, the baseline model is difficult to train without knowledge distillation. Even when initializing all non-Mamba weights from the teacher model, the GenEval score is still lower than 0.4. This justifies our claim that Mamba models are difficult to train, thus requiring knowledge distillation. We then add the proposed soft label loss, feature loss and layer-level teacher forcing to validate the effectiveness of knowledge distillation. The improvement in the two object score after adding layer-level teacher forcing component suggests that teacher forcing can help Mamba layers learn global context. It is noteworthy that with \ours{}, the image quality is greatly improved and even reaches the teacher model PixArt-$\alpha$, demonstrating the effectiveness of our method.

To study the intrinsic difference between causal and non-causal models, we add the result of causal-transformer-to-causal-Mamba initialization method~\cite{wang2024mamba} to non-causal transformer in DiT. Results show that directly initialzing the causal Mamba with non-causal can lead to degraded performance. This demonstrates that the formulation of causal Mamba and non-causal Transformer is fundamentally different.

\begin{table*}[t]
\centering
\resizebox{1.0\textwidth}{!}{
\begin{tabular}{c|cccccc|cc}
\shline
Method & position & counting & colors & attr\_binding & single\_object & two\_object & Overall$\uparrow$ \\
\hline
\hline
minDALL-E &    0.02   &   0.12   &   0.37   &   0.01    &    0.73     &    0.11       &  0.227 \\
SDv1.5~\cite{rombach2022high} &   0.04      &    0.35    &  0.76    &    0.06      &   0.97    &  0.38      &  0.427  \\
PixArt-$\alpha$~\cite{chen2023pixartalpha} \textit{(upper-bound)}  &    0.08    &  \textbf{0.48}     &  \textbf{0.81}  &      0.08  &     0.97     &     0.47      &  \underline{0.481} \\
\hline
\hline
baseline hybrid Mamba     &     0.02     &    0.18      &   0.60     &   0.04   &      0.08    &      0.18      &    0.301    \\
(+non-Mamba weight initialization)     &   0.05       &    0.27      &   0.71     &     0.06      &      0.94       &   0.37    &      0.397     \\
(+$\mathcal{L}_{soft}$)        &  0.05   &    0.42      &    0.79     &    0.08    &       \textbf{0.98}     &         0.41     &   0.452  \\
(+$\mathcal{L}_{feature}$) &   0.07       &   0.47       &  \textbf{0.81}      &     0.08       &       0.96         &     0.40      &      \textit{0.462}  \\
(+teacher forcing) [\textbf{T2MD}] & \textbf{0.10} & 0.44 & 0.79 & \textbf{0.14} & 0.96 & \textbf{0.49} & \textbf{0.485}  \\
\hline
\hline
Causal-to-causal Mamba initialization & 0.06 & 0.40 & 0.77 & 0.06 & 0.95 & 0.36 & 0.433 \\
Bi-dir $\rightarrow$ Uni-dir     &   0.06     &   0.44    &   0.77     &      0.07    &     0.96      &      0.39       &   0.448   & \\
No SA &    0.05    &   0.38   &   0.77     &   0.05       &  0.96    &   0.31     &   0.420 & \\
\shline
\end{tabular}
}
\caption{Ablations and quantitative comparisons under the GenEval benchmark. Teacher forcing and knowledge distillation are crucial for training Mamba models. \ours{} greatly improves the GenEval score by 0.184, and is comparable with the teacher model in GenEval score.}
\vspace{4px}
\label{tab:ablation}
\end{table*}

\subsubsection{Model adaptation}
The model adaptation stage allows the model to change a few components so that it can better adapt to the higher resolution generation. Surprisingly, even if we have replaced T5, VAE, and positional encodings, the model can still adapt to these new components within 100k training steps. As shown in Fig.~\ref{fig:512}, the adapted model quickly converges and learns to generate images correctly. 

\begin{figure}[t]
    \centering
    \includegraphics[width=1.0\linewidth]{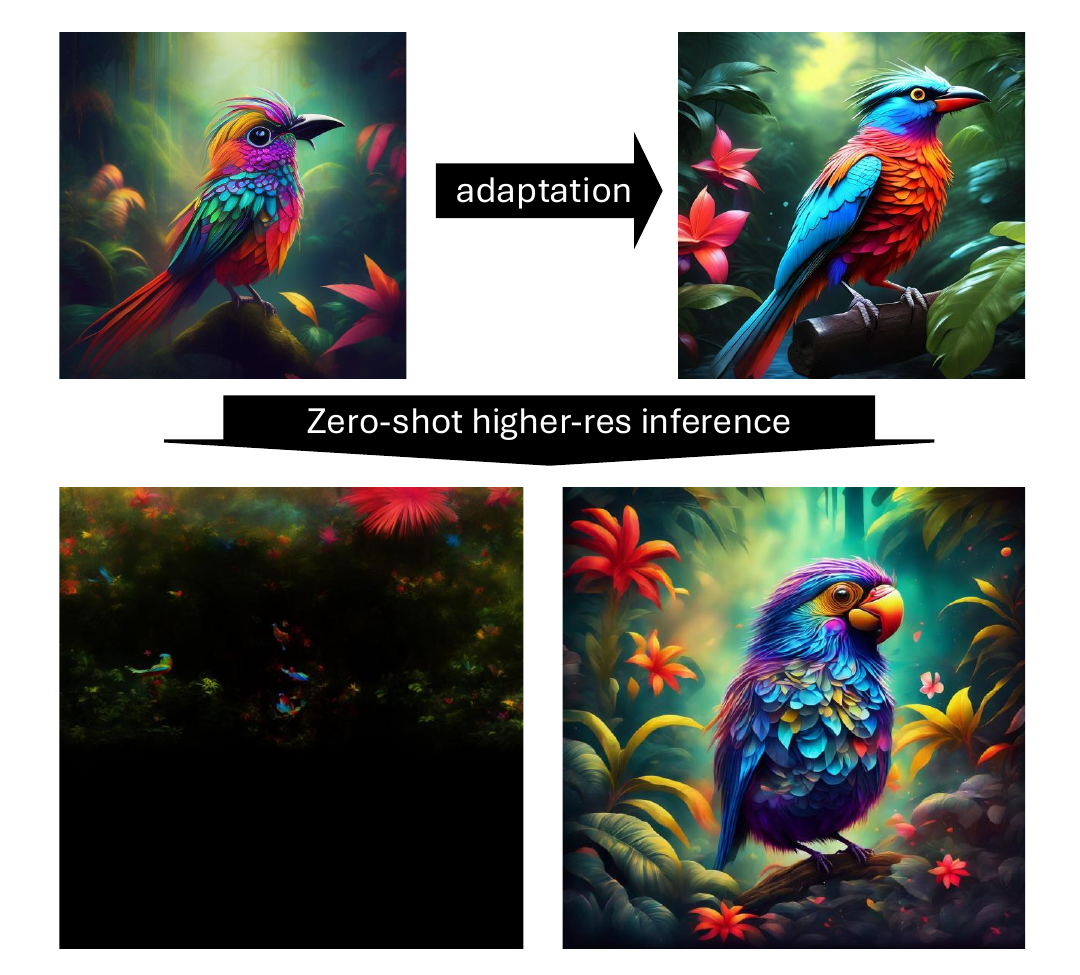}
    \vspace{-8px}
    \caption{Without model adaptation, the original 512$\times$512 student model is not capable of zero-shot 1024$\times$1024 generation.}
    \vspace{-14px}
    \label{fig:zero-shot}
\end{figure}

\begin{figure*}[t]
    \centering
    \includegraphics[width=1.0\linewidth]{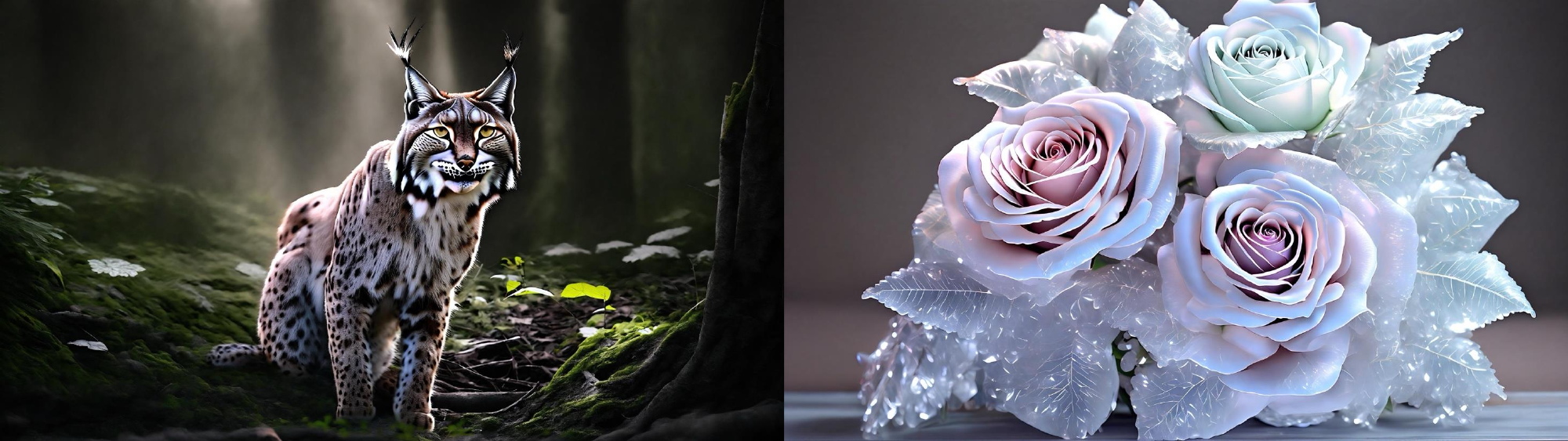}
    \caption{Examples of zero-shot 4k generation. Although our model has not been trained with 4k data, it is capable generate 4k examples. The input texts are \textit{Professional photograph of a lynx lit by moody harsh lighting in the middle of a forest.} and \textit{A bouquet of roses made of pastel ice crystals.}.}
    \label{fig:4k}
    \vspace{-8px}
\end{figure*}

Model adaptation enables the model to incorporate normalized positional encoding, thereby supporting zero-shot higher resolution generation. Fig.~\ref{fig:zero-shot} demonstrates the difference in zero-shot capabilities with or without model adaptation. Without adaptation, the previous model is not capable of zero-shot higher-resolution generation. The zero-shot higher-resolution generation capabilities of our model enable \ours{} to generate zero-shot 4K image results after the final stage. Results can be found in Fig.~\ref{fig:4k}.

\subsection{Comparisons with other methods}

\begin{table}[ht]
\centering
\begin{tabular}{l|ccc}
\shline
Method      & \#Params & FID-30K $\downarrow$ \\ \hline\hline
DALLE~\cite{ramesh2021zero}        & 12.0B    & 27.5    \\
GLIDE~\cite{nichol2021glide}         & 5.0B     & 12.24   \\
LDM~\cite{rombach2022high}            & 1.4B     & 12.64   \\
DALLE 2        & 6.5B     & 10.39   \\
StyleGAN-T~\cite{sauer2023stylegan}       & 1B       & 13.90   \\
SDv1.5~\cite{rombach2022high}      & 0.9B     & 9.62   \\
Dimba~\cite{fei2024dimba}    &  0.9B   &   8.93    \\
LinFusion~\cite{liu2024linfusion}     & 0.9B     & 12.57   \\
PixArt-$\alpha$~\cite{chen2023pixartalpha} \textit{(teacher)}   & 0.6B     & 7.32    \\ \hline \hline
T2MD (ours)     & 0.7B     & 8.63  \\
\shline
\end{tabular}
\caption{FID-30k results on the MS-COCO 2014 validation set (zero-shot). With only 0.7B parameters, T2MD achieves comparable or even better results than other methods.}
\vspace{-15px}
\label{tab:fid}
\end{table}

\subsubsection{Qualitative analysis}
In Fig.~\ref{fig:teaser}, we provide some of the results under 2048 $\times$ 2048 resolution and 2688$\times$1536 resolution.
We also present some 512$\times$512 distillation results in Fig.~\ref{fig:512}. From the visualizations, we can see that after knowledge distillation, our model is capable of generating high-fidelity images, and the image generation quality is comparable to the teacher model PixArt-$\alpha$.
This corresponds to our ablation results that teacher forcing and knowledge distillation are effective for training diffusion Mamba models.
Also, it is interesting to notice that in some cases where the teacher model doesn't align with the text prompts, the model can fix this issue after knowledge distillation, though the cross-attention layers are frozen throughout the knowledge distillation stage. This phenomenon demonstrates that our knowledge distillation training method allows the Mamba model to learn the semantic information contained within the model.

\subsubsection{Quantitative analysis}
We assessed the performance of our image generation model using the Frechet Inception Distance (FID)~\cite{heusel2017gans} under MS-COCO 2014 dataset~\cite{lin2014microsoft}.
We compared our method with the GAN-based method StyleGAN-T~\cite{sauer2023stylegan}, the widely-used diffusion model SD1.5~\cite{rombach2022high}, the knowledge-distillation based linear complexity model LinFusion~\cite{liu2024linfusion} and the distillation-based method InstaFlow~\cite{liu2023instaflow}. The results are presented in Tab.~\ref{tab:fid}. Despite having only 0.7B parameters, our proposed T2MD outperforms most other methods. This experiment demonstrates that our T2DM model can generate high-fidelity results with minimal number of parameters. The reason for our strong capability is that the Mamba-based T2DM model gains knowledge from the well-trained Transformer-based teacher model PixArt-$\alpha$. The result supports our initial claim that a sequential and casual state-space model is capable of modeling the long-range non-causal dependency among the visual tokens.

\subsection{Efficiency analysis}
To evaluate the efficiency of our model, we provide the efficiency analysis by demonstrating the throughput and latency on 1 H100 GPU. 
We compare the image sampling throughput and latency of the proposed \ours{}, a 0.7B hybrid diffusion Mamba model, against the 0.6B baseline DiT model.
The self-attention layers employing the Flash-Attention 2 implementation.
The speed comparison results are presented in Tab.~\ref{tab:speed}.
Our findings indicate that the hybrid architecture of \ours{} achieves substantial efficiency gains in high-resolution image synthesis. Notably, despite having 0.1B additional parameters, \ours{} is 1.5$\times$ faster than the DiT model for generating 2048$\times$2048 images in latency. This performance gap widens significantly in zero-shot 4K image generation, with \ours{} achieving a 2.1$\times$ speedup in latency. These results highlight the computational efficiency of our model, particularly in handling high-resolution data. The hybrid design of \ours{}, which combines the strengths of both transformer and diffusion architectures, enables it to process large-scale images with minimal latency.

\begin{table}[h]
\centering
\resizebox{1.0\linewidth}{!}{
\begin{tabular}{cllll}
\shline
resolution & throughput & latency & SA latency & speedup \\
\hline
2048$\times$2048 & 0.24 & 2.8s & 4.2s & 1.5$\times$ \\
3840$\times$2160 & 0.15 & 6.5s & 13.6s & 2.1$\times$ \\
\shline
\end{tabular}
}
\caption{Throughput is measured by \#samples per second. Latency is measured under bz 1.}
\vspace{-16px}
\label{tab:speed}
\end{table}
\section{Conclusion}
We present a novel approach to high-resolution image generation by leveraging a hybrid diffusion Mamba model. Through a multi-stage training process that includes layer-level teacher forcing, knowledge distillation, model adaptation, and high-resolution fine-tuning, \ours{} effectively distills knowledge from Transformer-based DiT models to achieve both computational efficiency and high-quality image outputs at up to 4K resolution.


\newpage
{
    \small
    \bibliographystyle{ieeenat_fullname}
    \bibliography{main}
}


\end{document}